# Deep Learning and Machine Vision for Food Processing: A Survey

Lili Zhu[a], Petros Spachos[a,*], Erica Pensini[a] and Konstantinos N. Plataniotis[b]

a School of Engineering, University of Guelph, Guelph, ON N1G 2W1, Canada
b Department of Electrical and Computer Engineering, University of Toronto, Toronto, ON M5S 3G4, Canada

**Abstract**

The quality and safety of food is an important issue to the whole society, since it is at the basis of human health, social development and stability. Ensuring food quality and safety is a complex process, and all stages of food processing must be considered, from cultivating, harvesting and storage to preparation and consumption. However, these processes are often labour-intensive. Nowadays, the development of machine vision can greatly assist researchers and industries in improving the efficiency of food processing. As a result, machine vision has been widely used in all aspects of food processing. At the same time, image processing is an important component of machine vision. Image processing can take advantage of machine learning and deep learning models to effectively identify the type and quality of food. Subsequently, follow-up design in the machine vision system can address tasks such as food grading, detecting locations of defective spots or foreign objects, and removing impurities. In this paper, we provide an overview on the traditional machine learning and deep learning methods, as well as the machine vision techniques that can be applied to the field of food processing. We present the current approaches and challenges, and the future trends.

## 1. Introduction

Food processing is the transformation of either raw materials such as natural animals and plants into food, or of one form of food into other forms which are more suitable for the dietary habits of modern people. Therefore, food processing is closely related to the quality of modern people's life and the economic development of the whole society. Food processing includes diverse aspects (e.g., the manufacturing, modification and manufacturing of food), which require a series of physical and chemical changes of the raw materials (Smith et al., 2011). However, due to an increase in environmental pollution, people are concerned about the safety of both the food sources and the food processing procedures. Throughout the process, it is necessary to ensure that the nutritional properties of the raw materials are retained, and that toxic and harmful substances are not introduced into the food. Therefore, food processing is of high value to food scientists, the food industry, and consumers (Augustin et al., 2016).

In order to meet society's increasing expectations and standards for food processing, the quality control of food and agricultural products must be accurate and efficient, and it has therefore become an arduous and labour-intensive task. The demand for food is increasing significantly because of the increase in the population, especially in some developing countries such as China and India (Bodirsky et al., 2015). After years of rapid development, machine vision has become common in diverse sectors, including agriculture, medical care, transportation and communications. Its high efficiency and accuracy can reduce labour costs, and even exceed human performance (Patel et al., 2012). In the agricultural field, agri-technology and precision agriculture are emerging interdisciplinary sciences that use machine vision methods to achieve high agricultural yields, while reducing environmental impact. Machine vision systems can acquire image data in a variety of land-based and aerial-based methods, and they can complete diverse tasks, including quality and safety inspection, agriculture produce grading, foreign object detection and crop monitoring (Chen et al., 2002). Specifically, in food processing, a machine vision system can collect a series of parameters, such as the size, weight, shape, texture and colour of the food, and even numerous details that cannot be observed by the human eye, with the goal of monitoring and control food processing. In this way, human errors caused by repetitive work are avoided (Cubero et al., 2010).

The rapid development of machine vision systems and image processing algorithms, as well as the increased number of food types and various food processing steps and methods, has led to a significant increase in the literature published on this subject matter. In this survey, we summarize representative papers published in this field in the last five years. These papers are organized based on the different functions and methods they used, to facilitate tracking the state-of-the-art methods of machine vision systems and image processing in the field of food processing.

The various notations and abbreviations which will be used in this survey are given in Table 1. A machine vision system is presented in Section 2, followed by a discussion on food processing in Section 3. In Section 4, the different approaches of machine learning are presented, while the open challenges and the future trends are in Section 5. The conclusion of the survey is in Section 6.

**Table 1** Abbreviations used in this survey.

| Abbreviation | Terms | Abbreviation | Terms |
|---|---|---|---|
| AA | Antioxidant Activity | MRI | Magnetic Resonance Imaging |
| AC | Ash Content | ms | millisecond |
| ANN | Artificial Neural Network | MSER | Maximally Stable Extremal Regions |
| AV | Anisidine Value | MVS | Machine Vision System |
| BN | Bayesian Network | NIR | Near Infrared |
| BP | Back Propagation | OCR | Optical Character Recognition |
| CCF | Comprehensive Color Feature | OCV | Optical Character Verification |
| CFS | Correlation based Feature Selection | PBR | Physically Based Rendering |
| CNN | Convolutional Neural Network | PCA | Principal Component Analysis |
| CV | Coefficient of Variation | PCR | Principal Component Regression |
| DSSAEs | Deep Stacked Sparse Auto-Encoders | PLS-DA | Partial Least Squares-Discriminant Analysis |
| DT | Decision Trees | PLSR | Partial Least Squares Regression |
| DWT | Discrete Wavelet Transform | PNN | Probabilistic Neural Network |
| ELMs | Extreme Learning Machines | PV | Peroxide Value |
| ExR | Excess Red | RGB | Red, Green, Blue |
| FA | Free Acidity | RID | Relative Internal Distance |
| FCA | Feature Color Areas | RMSE | Root-Mean-Square Error |
| FCM | Fuzzy C-means | ROI | Region of Interest |
| FCN | Fully Convolutional Network | RVM | Relevance Vector Machine |
| GDF | Gaussian Derivative Filtering | SIRI | Structured Illumination Reflectance Imaging |
| GLCM | Gray Level Co-occurrence Matrix | SLIC | Simple Linear Iterative Clustering |
| GLGCM | Gray Level-Gradient Co-occurrence Matrix | SMK | Sparse Multi-Kernel |
| GMM | Gaussian Mixture Model | SPA | Successive Projections Algorithm |
| HS | Histogram Statistics | SVM | Support Vector Machine |
| HSI | Hyperspectral Imaging | SVR | Support Vector Regression |
| ICA | Imperialist Competitive Algorithm | SWIR | Shortwave Infrared |
| KNN | K-Nearest Neighbors | TBARS | Thiobarbituric Acid Reactive Substances |
| LCTF | Liquid Crystal Tunable Filter | TOTOX | Total Oxidation |
| LDA | Linear Discriminant Analysis | TPC | Total Phenolic Content |

| LSSVM | Least Squares Support Vector Machine | UV | Ultraviolet |
|---|---|---|---|
| MADM | Multi Attribute Decision Making | UVE-SPA | Uninformative Variable Elimination and Successive Projections Algorithm |
| MD | Mahalanobis distance | VIS-NIR | Visible and Near Infrared |
| MLP | Multilayer Perceptron | YOLO | You Only Look Once |

## 2. Machine Vision System (MVS)

MVS are typically used in industrial or production environments, and can use one or several cameras to automatically observe, capture, evaluate, and recognize stationary or moving objects (Golnabi et al., 2007). The system then uses the obtained data to control the subsequent manufacturing procedures.

A MVS usually includes digital cameras, and image processing software running on computers and mechanical systems, as shown in Fig. 1. The illumination device provides sufficient light to the object so that the camera can capture high-quality images of the object. The software at the computer can then process the images for different purposes. The processed image results are used by the mechanical system to make the next operational decision.

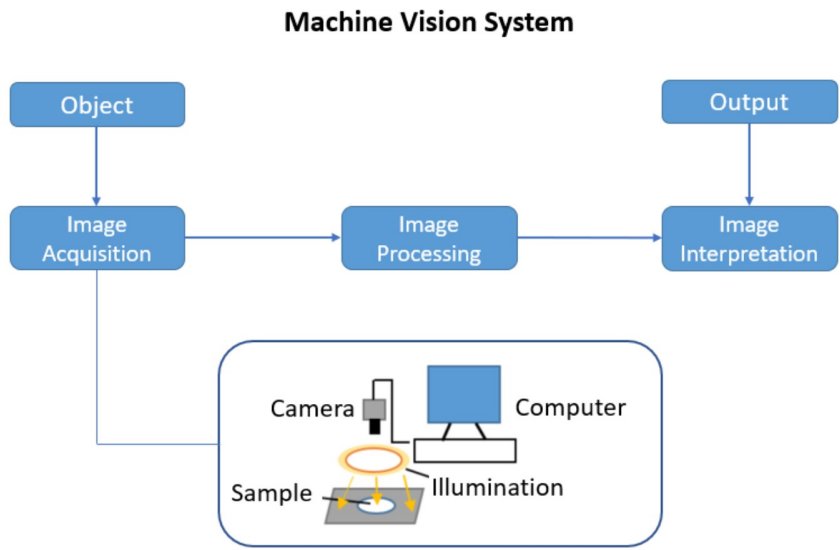

**Figure 1** Main components of a machine vision system.

**Table 2** Main processes of an MVS

| Step | Description | Methods |
|---|---|---|
| Image Acquisition | The process of obtaining scene images from the worksite and the first step in a machine vision system | Smart cameras, PC based systems, vision appliances, etc. |
| Image Processing | The technology of processing and analyzing images to meet visual, psychological or other requirements | Low level processing, intermediate processing, and high level processing. |

An MVS includes two main parts to enable objective and non-destructive food evaluation: 1) acquiring and 2) processing (Timmermans et al., 1995).

1. Image acquisition determines the quality and information of the images. It is the foundation of whether the subsequent image processing can be carried out effectively.

2. Image processing guides the operation of the machinery and has played a key role in the multi-tasking of machine vision systems (Du et al., 2004).

The main processes of a typical MVS are briefly described in Table 2. The rest of this section describes each part in detail.

## 2.1. Image Acquisition

An MVS can acquire images in real-time through photos, videos and other 3-dimensional (3D) technologies. The images can be transmitted to the processing unit by cable connection, Internet/Ethernet, radiofrequency identifications, and even wireless sensor networks (Bouzembrak et al., 2019). There are a number of ways to obtain high-quality images in the food processing field, as described in the following sub-sections.

### 2.1.1. Stereo systems

Stereo systems refer to a set of cameras that can measure the distance (depth) from an object to the camera. Compared with the traditional 2-dimensional (2D) camera, the stereo system increases the depth of one dimension, with the goal of improving its representation of the real world (Sing Bing Kang et al., 1995). With the increasingly widespread use of machine vision, it has become more common to use stereo systems to collect depth information of the environment and then perform object recognition and environment modelling. A stereo camera prototype is used to capture a pair of images, which are utilized to estimate the dimensions of the target food (Subhi et al., 2018). Other grading techniques based on depth images have also been proposed (Su et al., 2018). For instance, a depth camera was used to obtain 3D features of potatoes such as bumps, divots, and bent shapes. The length, width, and thickness information were processed to calculate the volume and the mass of the potatoes, based on which the grading task according to mass was achieved. Subsequently, the 2D and 3D surface data were integrated to detect the irregular shape, hump, and hollow defect of potatoes. The grading accuracy of the potatoes' appearance was 88%. The acquired information was also used to rebuild a virtual reality potato model, which may be useful for food packaging research.

### 2.1.2. Remote sensing image (RS)

RS refers to films or photos that record the magnitude of electromagnetic waves emitted by various objects and mainly includes aerial photos and satellite photos (Wang et al., 2009). Unlike normal images that only include red, green, and blue bands, RS includes multiple colour bands. Since objects retain their spectral features, different combinations of bands can be selected in RS according to different objectives. Visible light, Near Infrared (NIR), Thermal Infrared (TIR), Panchromatic and Shortwave Infrared (SWIR) are common spectral bands which are collected via sensors on satellites. Moreover, RS includes geographic information, which normal images do not have. Therefore, it is necessary to process the image geometrically. This process corrects the geographic information of the RS so that it can correspond to the coordinates of the ground target more accurately. Nowadays, RS is one of the essential techniques for precision agriculture (Seelan et al., 2003). Fan (2016) and Sun (2016) used near-infrared (NIR) spectroscopy to detect soluble solids content of apples and the apple firmness, respectively. Khatiwada (2016) adopted NIR spectroscopy to detect the internal flesh browning in intact apples.

### 2.1.3. Hyperspectral image

The hyperspectral image is composed of many channels and each channel captures the light of a specified wavelength. The data acquired by the hyperspectral device contains the image information and can also be expanded in the spectral dimension. As a result, it is possible to obtain both the spectral data of each point on the image and the image information of any spectrum. Hyperspectral images can be used in food safety, medical diagnosis, aerospace, and other fields. Sun (2017) presented a 360° rotating hyperspectral imaging system to detect fungi which cause peach decay at the peach surface. Moreover, hyperspectral reflectance imaging methods were applied to automatically detect the bruise or damage of blueberries (Hu et al., 2016; Fan et al., 2017). Fan (2016) and Tian (2017) combined spectra and textural features of hyperspectral reflectance imaging, respectively. Hyperspectral imaging was proposed as a tool to assess the quality of pear, strawberry, apricot, and pomegranate (Hu et al., 2017; Liu et al., 2014; Büyükcan et al., 2016; Khodabakhshian et al., 2016). Zhang (2015) adopted hyperspectral imaging and chemometrics combined method to detect egg freshness, egg internal bubbles, and egg scattered yolk.

### 2.1.4. X-ray

X-ray equipment generates X-rays and uses them to detect either metal foreign objects in food products or non-metallic foreign objects with higher density than the surrounding matrix. Therefore, X-ray technology offers a tool to detect contaminants in food. It can identify foreign objects such as metal, glass, calcified bone, stone, and high-density plastic, thereby ensuring food safety. Grating-based X-ray imaging was used to acquire images of food products and detect foreign objects in food products (Einarsdóttir et al., 2016). Purohit et al. adopted X-ray diffraction to analyze the quality of certain food such as grains, pulses, oilseeds, and derived products (Purohit et al., 2019).

### 2.1.5. Thermal imaging

Thermal imaging is a technique that detects infrared energy (heat) without contact and converts it into an electrical signal, to generate a thermal image of objects and determine their temperature. Thermal imaging has various applications in the food industry (Vadivambal et al., 2011). It can monitor and control the temperature of food during pre-treatment, pre-heating, and disinfection. It can also detect damage to fresh produce, as was for instance demonstrated for guava fruit, under different temperatures conditions where damage to guava was detected based on infrared (IR) thermography data (Gonçalves et al., 2016).

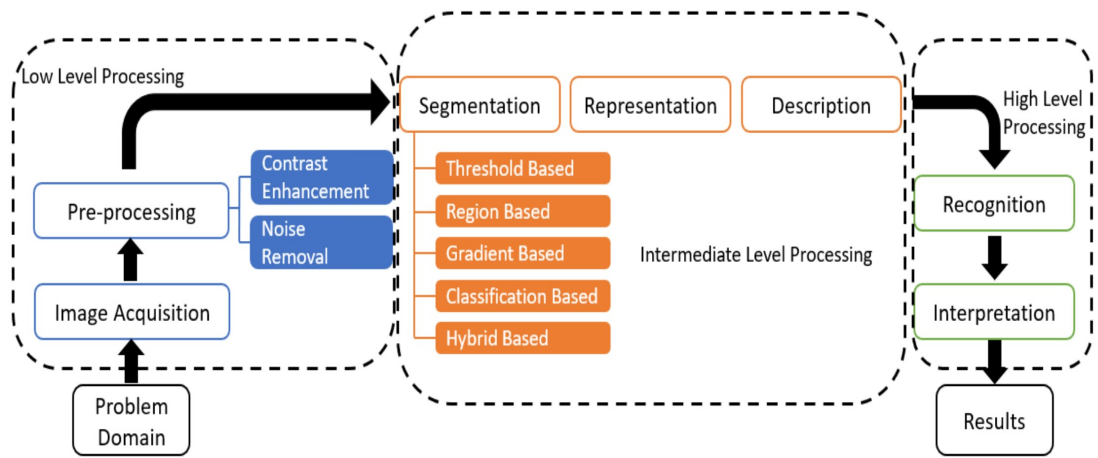

**Figure 2** Different levels in image processing process.

### 2.1.6. Magnetic resonance imaging (MRI)

MRI is a non-destructive imaging method based on nuclear magnetic resonance (Ebrahimnejad et al., 2018). This method enables determining the characteristics of objects of interest (including food), by tracking proton motion within them with good spatial resolution (Ebrahimnejad et al., 2018). Since many raw and processed foods are heterogeneous systems, traditional research methods cannot estimate heat transfer, moisture migration, and temperature distribution within the food (Ebrahimnejad et al., 2018). MRI has also been successfully used to determine the texture of foods, such as pork meat (Caballero et al., 2017).

### 2.1.7. Additional imaging methods

Some other imaging methods can also help acquire the information of the target objects. For instance, Yu (2015) developed an orthogonally polarized Terahertz (THz)-wave imaging system (operating at 0.3 THz) to inspect food. The system developed could generate images of foods moving on a moving conveyor belt and detect foreign objects. Vasefi (2018) presented a multimode tabletop system and adopted spectroscopic technologies for food safety and quality applications. Gómez-Sanchís et al. (2014) designed a hemispherical illumination chamber to illuminate spherical samples, and developed a liquid crystal tunable filter (LCTF) based method to image spherical fruits. The authors also used segmented hyperspectral fruit images to detect fruit ripeness with >98% accuracy (Gómez-Sanchís et al. (2014)).

## 2.2. Image Processing

Image processing is used to generate new images based on existing images, with the goal of extracting or improving the characterization of a region of interest (ROI). This process can be seen as digital signal processing and does not involve interpreting the content or meaning of the image. Image processing contains low level processing, intermediate level processing, and high level processing. The different levels of the image processing process are shown in Fig. 2 (Brosnan et al., 2004).

### 2.2.1. Low level processing

Low level processing is used to pre-process images (Senni et al., 2014). When imaging objects, different imaging and sensing devices can be simultaneously used to acquire images of samples and convert them into digital forms, which can be read by a computer. The images acquired can be imperfect because of diverse reasons, including insufficient lighting, long-distance or low resolution of the imaging device, unstable viewfinder, and other interferences. Therefore, it is usually necessary to pre-process the original image to improve the final image analysis. Common image pre-processing methods include image enhancement, such as adjusting the brightness or colours of the images, cropping the images to focus on the ROI, and removing noise or digital artifacts from low light levels. Senni (2014) proposed an on-line Thermography Non-Destructive Testing method to detect whether biscuits were contaminated with foreign objects. An off-line step was designed to analyze the decay curves of foreign objects and biscuits with an infrared camera positioned at the exit of the oven when the biscuits were cooling down. Later an online system was installed with the Thermo camera located at a 75cm distance above the conveyor belt, which had a 3m distance from the oven exit. During the image pre-processing step, the authors applied a 2D low pass-filter, a focusing filter, and a 2D Wiener filter in the frequency domain to reduce the Additive White Gaussian Noise, image blurring, and drag effects problems in the raw images.

### 2.2.2. Intermediate level processing

Intermediate level processing includes image segmentation, image representation, and image description (Shih, 2010). Image segmentation is one of the most essential steps in image processing, since it largely determines whether image analysis focuses on the target sample. The function of image segmentation is to separate the target from other unwanted image information, thereby reducing the computational cost of subsequent image analysis, while also improving accuracy. Watershed segmentation is an image segmentation algorithm that is based on the analysis of geographic morphology. It imitates geographic features (such as mountains, ravines, and basins) to classify different objects. Li (2018) provided a fresh improved watershed segmentation method based on morphological gradient reconstruction and marker extraction to segment bruised regions on peaches. Ni (2018) created an apple-picking robot vision system to estimate the position and location of apples after segmenting apples via the Otsu's method in a complex background. Momin (2017) proposed an MVS to grade mango fruits. In this system, the images of mangoes were captured by a colour camera and image segmentation techniques including thresholding, pattern recognition, and deformable models were used to perform the segmentation. Subsequently, features such as projected area, perimeter, Feret diameter, and roundness of the fruits were identified to assess the geometrical and shape properties of the fruit. As a final step, a simple filtering operation was used to count pixels in projected areas to grade the mangoes into three small, medium, and large categories.

Pan (2016) developed a sliding comparison local segmentation algorithm to detect the defects on oranges and recognize their stem ends. The proposed system was used image 1191 oranges, with 93.8% accuracy in distinguishing stem ends from orange peels and 97% accuracy in detecting defective samples. Image representation includes the boundary representation and the regional representation (Shih, 2010). The boundary representation is used to describe the size and shape characteristics, while the regional representation is applied to describe the texture and defects of the image (Shih, 2010). Image descriptions can extract quantitative information from images that have been processed in previous steps. Hosseinpour (2015) designed a novel rotation-invariant and scale-invariant image texture processing approach, to eliminate the effects of sample shrinkage on the detection the texture features during the monitoring of in-line shrimp drying. Image of texture features were fed into a Multilayer Perceptron-ANN (MLP-ANN), to predict the moisture level of shrimps.

### 2.2.3. High level processing

High level processing includes image recognition and image interpretation. In this step, statistical methods or deep learning methods are commonly used to classify the target, based on the application of interest. The results of the analysis determine the processing required in the subsequent steps. Algorithms such as K-nearest neighbour (KNN), Support Vector Machine (SVM), neural network, fuzzy logic, and a genetic algorithm can help interpret the information obtained from the image. Neural networks and fuzzy logic methods have been successfully applied to MVS in the food industry. Kashyapa (2016) proposed a machine vision-based system to remove the calyx from strawberries. All strawberries on the conveyor were first photographed and then were oriented and positioned correctly, enabling calyx removal using waterjets, before strawberries were either further processed or packaged. Misimi (2016) developed a robot called "GRIBBOT" to move chicken fillets. The robot was equipped with a gripper resembling the curved beak of vultures, called "gribb" in Northern Europe (hence the robot's name). A robot with a multi-functional gripper tool was developed to realize the objective. An image processing algorithm was designed to locate the chicken fillets and the robot provided the 3D coordinates to its gripper. The average accuracy in pixels of the camera calibration was 0.2468 (root mean squared error, RMSE). Siswantoro (2013) predicted the volume of an irregular shape food product, by extracting the silhouettes of the targets from five different aspects and utilizing the Monte Carlo method to compute the volume of the target. The author used a bounding box with a real-world coordinate system to cover the object, and subsequently generated 3D random points in the bounding box. The Monte Carlo method was used to calculate the volume via the coordinates of those 3D points.

Image interpretation techniques commonly used on different elements of images are shown in Fig. 3, while Table 3 summarizes the non-machine learning classification methods mentioned in this section.

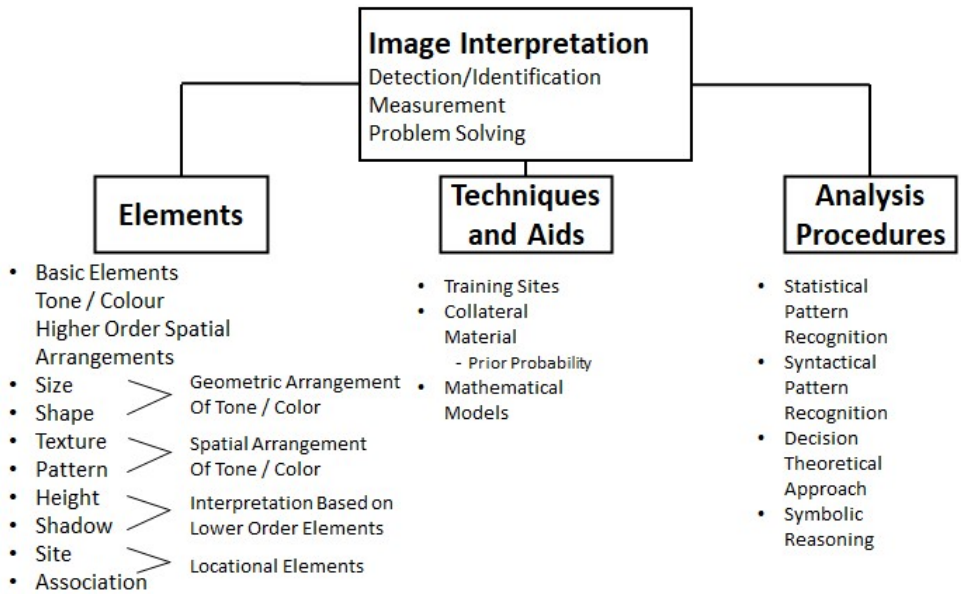

**Figure 3** Methods of Image Interpretation.

**Table 3**. Non-ML classification methods for food processing

| Products | Species | Application | Classification Methods | Evaluation | Reference |
|---|---|---|---|---|---|
| Fruit | Apple | Locating | Otsu | N/A | (Ni et al., 2018) |
| | Mango | Grading | count pixels | accuracy = 97% for mass, 79% for perimeter, and 36% for roundness | (Momin et al., 2017) |
| | Orange | Classifying | Sliding comparison local segmentation algorithm | accuracy = 97% | (Pan et al., 2016a) |
| | Strawberry | Monitoring | Self-improved algorithm | N/A | (Kashyapa, 2016) |
| Animal | Chicken | Monitoring | GRIBBOT | RMSE = 0.2468 | (Misimi et al., 2016) |
| Others | Biscuit | Detecting | thresholding algorithm | N/A | (Senni et al., 2014) |
| | Egg | Grading | hyperspectral imaging and chemometrics combined method | N/A | (Zhang et al., 2015) |
| | Irregular shape food | Monitoring | Monte Carlo method | mean absolute relative error 0.36%, and CV 0.34% | (Siswantoro et al., 2013) |

## 3. Food Processing

Food processing converts raw materials into new edible materials with improved properties (Fellows, 2009). Food processing is classified as primary and deep, depending on the degree of processing. Primary processing (rough processing) refers to the initial processing of agricultural products after harvesting, to keep the original nutrients of the product from being lost or to meet the requirements of transportation, storage, and reprocessing. The process principles and processing technology of primary processing are simple, and the commodity value is low. Examples of primary processing include drying, shelling, milling of grains, slaughtering of live animals and poultry, freezing processing of meat, eggs, and fish. Deep processing (finishing) refers to more elaborate processing procedures, used to further improve the characteristics of products, following primary processing. For example, grains can be processed into bread, noodles, biscuits, vermicelli, or soy sauce. Deep processing is an important way to increase the economic value of agricultural products (Fellows, 2009).

Most operations in traditional food processing are labour-intensive. The increased human population and diversification of consumer demands has created significant pressure on the food industry. This is because labour-intensive operations do not enable optimizing resources, thereby leading to high labor costs and even waste of raw materials, which further increases costs. Therefore, this food processing model cannot simultaneously yield high quality and low price (Capitanio et al., 2010). Although traditional methods are still playing an important role in food processing, food industry practitioners and researchers keep working on applying innovative and emerging food processing techniques to assist food processing to reduce costs, enhance the quality of food and improve processing efficiency (Van Der Goot et al., 2016). State-of-the-art technology can be used in almost every single step in the food supply chain, from farms to factories, and then to consumers (Manzini et al., 2013). For example, new technologies of MVS can be integrated into the procedures of freezing, drying, and canning for advancing the preservation period of food, and can be utilized to packaging and detecting foreign objects for increasing the shelf-life of food (Langelaan et al., 2013).

During the processing of raw food materials, some auxiliary procedures are needed to identify high-quality and sub-standard foods. These procedures include food safety and quality evaluation, food processing monitoring and packaging, and foreign object detection. In this way, it is possible to effectively prevent the waste of resources and improve productivity. Appearance, texture, and components of food are the primary references for determining the quality and safety of food, monitoring food processing procedures, and detecting foreign objects (Cardello et al., 2007; Ma et al., 2008). Image processing techniques are desirable options for the acquisition of this information. Moreover, the results of image processing can lead the system to determine the next steps about how to deal with different categories of food. These considerations justify the use of MVS in food processing. Here we will focus on the use of MVS in the context of food safety and quality evaluation, food process monitoring and packaging, and foreign object detection.

### 3.1. Food Safety and Quality Evaluation

All along, food safety and quality evaluation have attracted widespread attention worldwide. Food safety is vital to people's lives, ensuring people's health and providing a basic guarantee for the stable development of society (Yan, 2012). Food safety issues such as food contamination and deterioration are not acceptable by both the government and the public (Organization et al., 2003).

Food safety involves ensuring that food does not contain any toxic substances and that it meets the required nutritional requirements. Ensuring food safety requires an interdisciplinary approach, to ensure good hygiene during food processing, storage, and sale, with the goal of mitigating the risk of biotic and abiotic contaminants, which would lead to food poisoning.

Traditional food safety and quality evaluation methods are often impractical, because they are time-consuming and destructive. The food industry and consumers urgently need rapid, non-destructive ways to assess the safety and quality of food.

The morphological features of food (e.g., size and shape) are commonly used to grade food. This is because products having unusual or uneven shapes will affect their popularity in the market (Cubero et al., 2011). In the food industry, the size, shape, colour, and texture of fruits, vegetables, meat, and poultry are related to the price of the commodities. Furthermore, physical changes in the appearance of food can reflect changes in its intrinsic chemical state, especially when the food is about to deteriorate. Therefore, classifying food into different categories requires high-throughput, accurate methods. MVS are able to observe undesirable changes to identify foods that do not meet the specifications, thereby eliminating foods that are not suitable for the market.

The size and shape of food can be quantitatively characterized with image processing techniques, for instance by measuring the projected area and the perimeter. Naik (2017) proposed a system which used ripeness and size as key features to grade "langdo", a mango variety. The degree of mango ripeness is based on the mean intensity algorithm in L*a*b* colour space (where L* stands for perceptual lightness, and a* and b* for the chromaticity coordinates), while the mango size is predicted using a Fuzzy classifier with parameters of weight, eccentricity, and area.

Moreover, colour and texture features are often used to evaluate the quality and safety of meat and poultry. For example, the variation of protein, moisture, or fat in meat is likely to cause variations of the colour, water content, and texture of meat. Therefore, features like colour and texture can be used to determine the safety and quality of meat (Zapotoczny et al., 2016). Different algorithms can be used to determine such features, such as Local Binary Pattern, Gray Level Co-occurrence Matrix, and Gabor filter.

## 3.2. Packaging Monitoring

Modern food processing procedures include a number of complex steps that require strict control, adhering to regulations and ensuring both food quality and the safety of workers.

The first step is food processing, which often involves subjecting food to temperature and pressure changes that can affect the state of foods. Traditional control methods require staff to observe changes in food at all times, leading to human error (Linko et al., 1998). The automation and remote monitoring of food processing are key to guarantee quality.

After food processing, food packaging is used to transport and store food, safeguarding its quality. Effective, high quality packaging delays food decay, improving the efficiency of food distribution and marketing.

Food packaging has four essential functions: sanitation, food protection, product display, and ease of transportation (Han, 2005). Food packaging prevents food from being damaged due to the biological, chemical, and physical factors during the whole food supply chain. It can also maintain the stable quality of the food. The package's visual impact can show customers the appearance of food and offer them a first impression of the product. Therefore, food packaging is an indispensable component of the food manufacturing system. Similar to the monitoring of food processing, non-automated monitoring of food packaging can lead to human error and low efficiency (e.g., mixing undesirable objects into the package). These shortcomings can be addressed by using novel technologies, such as a machine vision system, to monitor food processing and packaging.

## 3.3. Foreign Object Detection

Foreign objects contamination is one of the main reasons for food recalls and rejection by consumers (Soon et al., 2020) This type of contamination harms customers and leads to loss of brand loyalty, in addition to causing large recall costs. Examples of foreign objects are insects, glass, metal, or rubber. These objects may inadvertently enter the food or package during any step of food processing. For example, small gravel or stone, insects, and twigs may enter at any step of harvesting, processing, handling, or preparation. Although the degree of danger of foreign objects depends on the size, type, hardness and clarity of the object, the ingestion of food containing foreign objects can cause choking or other diseases (Mohd Khairi et al., 2018).

As the food industry continues to improve the safety and quality of food products, the frequency of foreign objects in food is decreasing. At the same time, consumers are less accustomed to carefully checking for the presence of foreign objects in food and less prone to tolerating them. Detection of foreign objects with the naked eye is difficult. In contrast, emerging technologies and recognition methods can now easily detect foreign objects. Examples of such technologies include X-ray, ultrasound, thermal imaging (Mohd Khairi et al., 2018). For example, the acoustic impedance difference between meat and foreign objects enables foreign object detection in meat using ultrasound imaging (Xiong et al., 2017). This technique is precise and non-destructive, rendering its use promising in the food industry.

## 4. Machine learning approaches

Machine learning approaches use training samples, to subsequently discover patterns, and ultimately achieve an accurate prediction of future data or trends (Alpaydin, 2020). With the development of machine learning approaches and the increasing scale and difficulty of data analysis, traditional machine learning methods have often become unsuitable. As a result, deep learning methods have been developed, because they have more complex architectures and improved data analyzing capacity. The relationship between artificial intelligence, machine learning, and deep learning is illustrated in Fig. 4. In the following subsections, the algorithms and applications based on traditional machine learning methods and deep learning methods will be introduced separately.

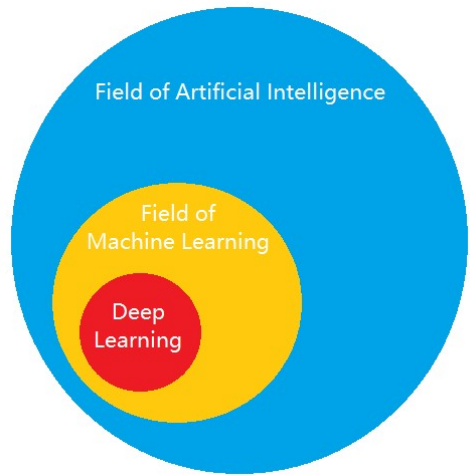

**Figure 4** The relationship between artificial intelligence, machine learning, and deep learning.

### 4.1. Traditional Machine Learning Methods

Traditional machine learning involves manually extracting features on small sample sets, to balance the validity of the learning results with the interpretability of the learning model, and to provide a framework for solving the learning problem when limited samples are available. Traditional machine learning can be sub-divided into supervised learning, unsupervised learning, and reinforcement learning, based on the training method and depending on whether the training data have labels or not. One of the important theoretical foundations of traditional machine learning methods is statistics, and the main analysis includes pattern classification, regression analysis, probability density estimation or other analyses. Related algorithms include Support Vector Machine, Logistic Regression, K-nearest neighbours, K-means, Bayesian Network, Fuzzy C-means, and Decision Tree.

#### *4.1.1. Algorithms*

***Support Vector Machine (SVM).*** SVM belongs to supervised learning and classifies data by finding a hyperplane that meets the classification requirements and renders the samples in the training set as far from the hyperplane as possible (Cortes et al., 1995). Based on this principle, the classification task is converted to a convex quadratic programming problem that requires maximizing the interval, and it is also equivalent to the minimization problem of a regularized hinge loss function (Fig. 5(a)). However, the data which need to be processed are not always linearly separable. Consequently, it is arduous to find the hyperplane that satisfies the condition. Therefore, the approach is to choose a kernel function for the SVM to solve nonlinear problems. The kernel function is applied to map the input in the low-dimensional space to the high-dimensional space. Subsequently, the optimal separating hyperplane is created in the high-dimensional space to separate the nonlinear data, as shown in Fig. 5(b).

***Logistic Regression.*** Logistic Regression is a machine learning method used to solve binary classification (0 or 1) problems and to estimate the probability of a certain scenario (Kleinbaum et al., 2002). The result of logistic regression is not the probability value in the mathematical definition, and it cannot be used directly as a probability value. The result is the ratio of the probability of the predicted category occurring and not occurring, which is called odds. As a result, the category can determined, which is helpful for tasks that use probability to assist decision-making. For example, it can be used to predict if a certain customer will buy a specific product or not.

***K-Nearest Neighbours (KNN).*** KNN classifies by measuring the distance between different feature values. The idea is that if most of the K similar samples in the feature space (the closest neighbours in the feature space) belong to a certain category, then the sample also belongs to this category. K is usually an integer not greater than 20 (Altman, 1992). In the KNN algorithm, the selected neighbours are all objects that have been correctly classified. This method only decides the category to which the samples to be classified belong, based on the category of the nearest sample or samples in the classification decision. In KNN, the distance between objects is calculated as a non-similarity index between objects, to avoid the matching problem between objects. The distance is usually the Euclidean or Manhattan distance (defined in (1) and (2), respectively). KNN algorithm is illustrated in Fig. 6, where it also shows that the result of the KNN algorithm largely depends on the choice of the K value.

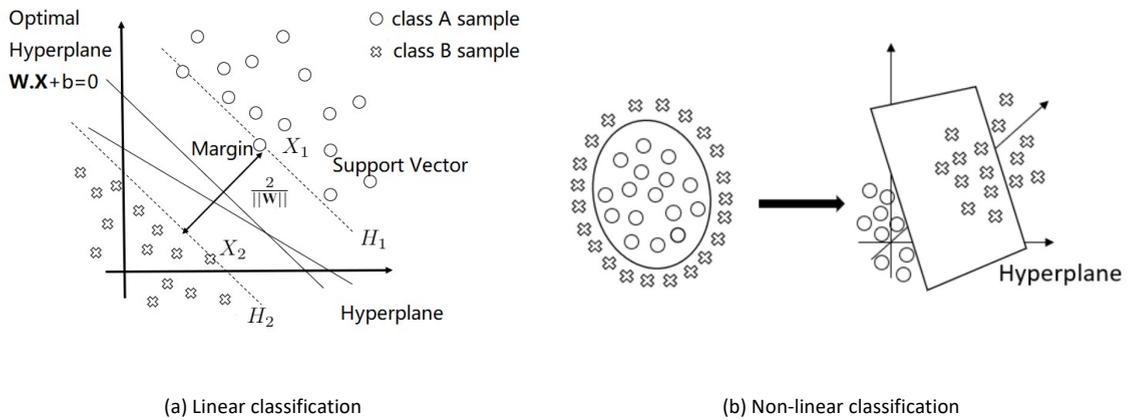

(a) Linear classification  (b) Non-linear classification

**Figure 5** Data classification with the SVM.

$$Euclidean\ distance:\ d(x,y) = \sqrt{\sum_{k=1}^{n}(x_k - y_k)^2} \qquad (1)$$

$$Manhattan\ distance:\ d(x,y) = \sqrt{\sum_{k=1}^{n}|x_k - y_k|} \qquad (2)$$

***K-means*** K-means is one of the most commonly used unsupervised learning methods in clustering algorithms. The clustering algorithm refers to a method of automatically dividing a group of unlabeled data into several categories. This method needs to ensure that the data of the same category have similar characteristics. However, it can only be applied to continuous data and requires the manual specification of the number of categories before clustering.

The similarity between data can be measured using Euclidean distance, which is an important assumption of K-means (Likas et al., 2003). A K-means clustering example of the same data set but with different K values is shown in Fig. 7.

***Bayesian Network.*** Bayesian network is a model of causal reasoning under uncertainty based on the Bayesian method. Mathematically, let $G = (I, E)$ represent a directed acyclic graph (where $I$ is the set of all nodes in the graph and $E$ is the set of directed connected line segments) and let $X = (X_i), i \in I$ be the random variable that is represented by a node $i$ in the directed acyclic graph (Friedman et al., 1997). If the joint probability of node $X$ can be expressed as shown in (3):

$$p(x) = \prod_{i \in I} p(x_{\{i\}} | x_{\{pa(i)\}}) \tag{3}$$

then $X$ is a Bayesian network relative to a directed acyclic graph $G$, where $pa(i)$ are the parents of node $i$. Moreover, for any random variable, the joint probability can be obtained by multiplying the respective local conditional probability distributions (expressed in (4)):

$$p(x_1, \ldots, x_K) = p(x_K | x_1, \ldots, x_{K-1}) \ldots p(x_2 | x_1) p(x_1) \tag{4}$$

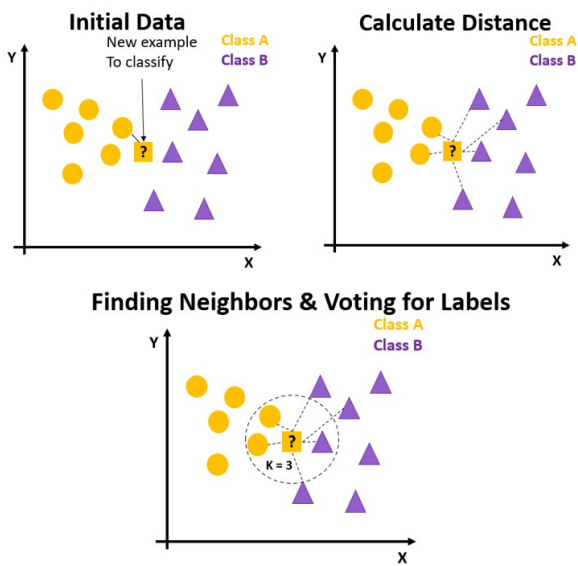

**Figure 6** An example of KNN algorithm.

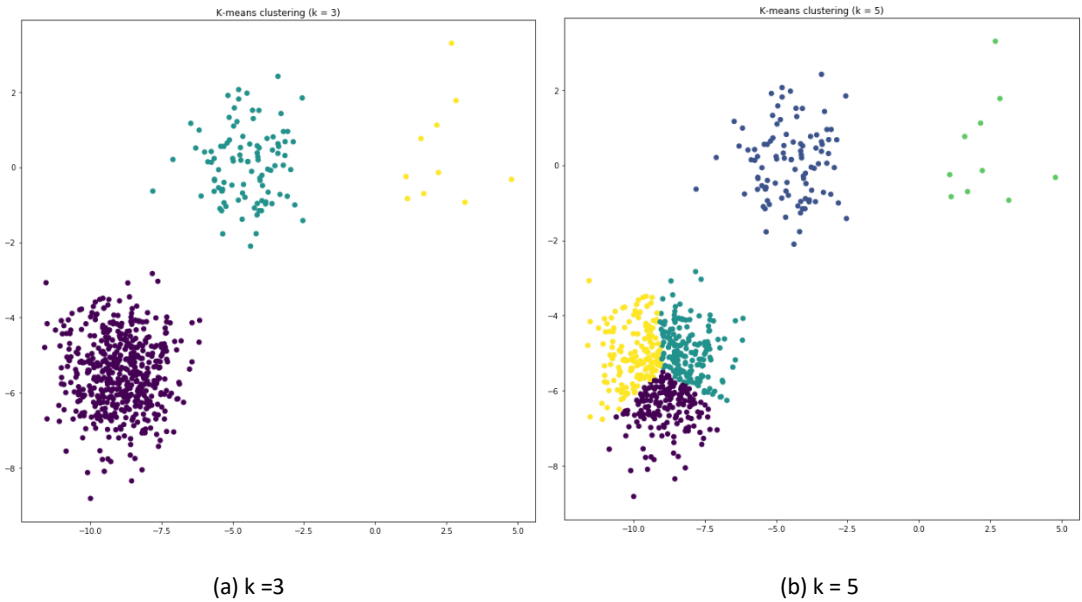

(a) k =3    (b) k = 5

**Figure 7** A K-means clustering example (Created with Python 3.5).

***Decision Tree.*** The decision tree is a tree structure in which each internal node represents a judgment on an attribute, each branch represents the output of a judgment result, and each leaf node represents a classification result (Safavian et al., 1991). The logic of the decision tree is shown in Fig. 8.

***Random Forests (RF).*** RF consists of different decision trees, and there is no correlation between these decision trees. When performing a classification task, a new input sample will enter, and each decision tree in the forest will independently determine its decision. The decision that appears most frequently among all classification results will be considered the final result (Breiman, 2001). An example of the relationship between the decision trees and random forests is shown in Fig. 9.

***Fuzzy C-means (FCM).*** Unlike the K-means algorithm (which clusters each sample into a certain class and assigns a label to the sample), fuzzy clustering assigns a probability vector (rather than a label) to a sample (Ghosh et al., 2013). This vector represents the probability that the sample belongs to a certain category. The FCM algorithm is an unsupervised fuzzy clustering method based on the optimization of the objective function.

***Dimensionality reduction methods.*** In practical applications, when the number of features increases beyond a certain critical point, the performance of the classifier decreases. This issue is also referred to as the "curse of dimensionality". For high-dimensional data, the "curse of dimensionality" makes it difficult to solve the pattern recognition problem. As a result, it is often required to reduce the dimension of the feature vector at first. Principal Components Analysis (PCA) and Linear Discriminant Analysis (LDA) are widely used dimensionality reduction methods for feature selection. PCA and LDA have many similarities. Both these methods map the original sample in a sample space with a lower dimension, although the mapping goals of PCA and LDA are different. PCA maximizes the divergence of the mapped sample, while LDA maximizes the classification performance of the mapped samples (Van Der Maaten et al., 2009). Therefore, PCA is an unsupervised dimensionality reduction method, and LDA is a supervised dimensionality reduction method.

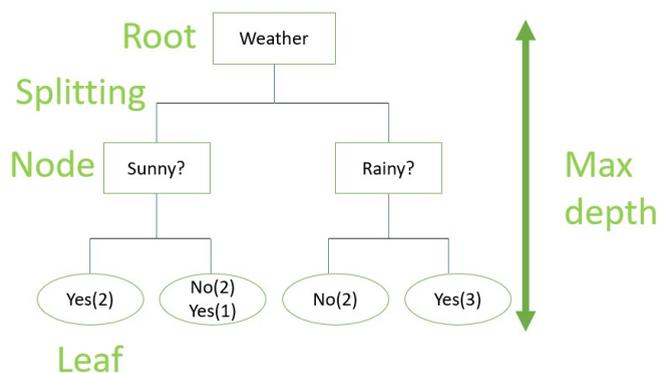

**Figure 8** An example of Decision Tree.

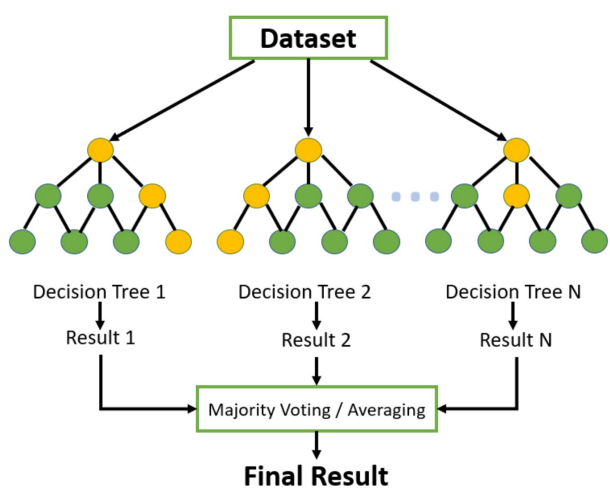

**Figure 9** An example of Random Forests.

### *4.1.2. Traditional ML in Food Safety and Quality Evaluation*

Leiva - Valenzuela (2013) proposed MVSs for the automatic detection of defects in blueberries. They used a pattern recognition algorithm to separate stem and calyx, and detected blueberries with diseases as well as the orientation of the blueberries. The authors then used LDA, Mahalanobis distance (data covariance distance), KNN (K was fixed to 5), and SVM as testing models to determine the optimal classifier.

Nandi (2016) graded mangoes using Multi-attribute Decision Making (MADM) and successfully predicted when the harvested mangoes could be shipped to market using Support Vector Regression (SVR). Finally, the authors used fuzzy incremental learning algorithm to evaluate the grade of mangoes, based on SVR and MADM. MADM refers to decision-making in a limited (infinite) set of solutions that conflict with each other and cannot be shared. Amatya (2015) developed an application for the automated cherry harvesting with an MVS. This application could be used to predict which branches were partially covered by foliage, using a Bayesian classifier which achieved an 89.6% accuracy in branch pixel classification.

Keresztes (2016) proposed an early apple bruise detection system, which included a SWIR illumination unit and a line scan camera. The authors used thirty Jonagold apples to demonstrate the effectiveness of their detection system in identifying fresh bruises. The image pixels were sub-divided into three sub-categories: 1) bruised, 2) unbruised, and 3) glare. Significant differences in the reflectance values between the three classes were observed. The best classifier to pre-process the images was the partial least squares discriminant analysis (PLS-DA) classifier. PLS-DA can reduce the impact of multicollinearity between variables. The post-process method used was physically based rendering (PBR), which means that the entire rendering algorithm or process can achieve physical reality, with mean centering as the reluctance calibration. The prediction of bruised apples achieved 98% accuracy. The processing time was 400, 300, and 200 milliseconds (ms) for the three increasing speeds of 0.1, 0.2, and 0.3 m/s, which were rapid enough for a real-time conveyor to process the apple sorting.

Zhang (2014) proposed a system with automatic lightness correction and weighted relevance vector machine (RVM) to detect defects on apples, and the overall accuracy of the prediction was 95.63%.

Extreme Learning Machines (ELM) uses random numbers and the law of large numbers to solve problems. Iraji (2018) combined multiple input features, neural networks, regression, and ELM to a system named Multi-layer Architecture of a SUB-Adaptive Neuro-Fuzzy Inference System. The system and a Deep Stacked Sparse Auto-Encoders (DSSAEs) were used to grade tomatoes, using the full tomato images (rather than pre-processed images from which specific features were extracted). The accuracy of the novel method developed by the authors was 95.5%. Noordam (2000) installed a 3-CCD line-scan camera and mirrors to capture a full view of potatoes. The author applied a method that combined LDA and a Mahalanobis distance classifier to segment pixels in the images, to detect the appearance defects of potatoes. This study also used a Fourier-based shape classification method to detect misshapen potatoes. The proposed MVS was could effectively grade and inspect the quality of either red or yellow skin-coloured potatoes. The processing capacity of this system is about 50 potatoes per second. Su (2018) demonstrated another grading technique based on depth images. Theis study used a depth camera to obtain 3D features of potatoes such as bumps, divots, and bent shapes. Information regarding the length, width, and thickness was processed to calculate the volume and the mass of the potatoes, and finally grade them. Moreover, the 2D and 3D surface data were integrated to detect the irregular shape, hump, and hollow defect of potatoes. The appearance grading accuracy was 88%. The acquired information was also used to rebuild a virtual reality potato model, which may be useful for food packaging research.

Wang (2015) designed a LabVIEW program to obtain colour, spectral, depth, and X-ray images of onions, to assess both the internal and external quality of onions. This research also proved that an ensembled SVM classification accuracy and a single SVM classifier training with all multimodality features was higher than a single SVM classifier training with single-mode features. He (2014) proposed a method to acquire the images of fresh farmed salmon fillets with a hyperspectral imaging system. Information in the VIS-NIR (Visible and Near Infra-Red) spectral region was extracted, to predict non-destructively the pH and drip loss attributes in the fillets. Amani (2015) utilized image processing methods to measure the total amount of volatile nitrogen in beef stored at 4°$C$ for 0, 4, 8, and 12 days, and built a co-relation between the total volatile nitrogen level and the colorimetric parameters. The authors built a regression model to assist in the prediction of beef decay level. Xiong (2015) proposed a 400 - 1000 nm spectral range VIS-NIR hyperspectral imaging (HSI) system to conduct fast and non-destructive detection of thiobarbituric acid reactive substances (TBARS) in chicken meat. An imaging algorithm was developed to transfer the wavelengths selected based on a successive projection algorithm and partial least square regression (SPA– PLSR) model to each pixel in the image, with the goal of developing distribution maps. Yang (2018) integrated spectra and texture features via hyperspectral imaging at 1000 - 2500 nm range and adopted the PLSR model to predict the water-holding capacity of chicken breast fillets.

Ramos (2017) designed an MVS to count and find harvestable and not harvestable fruits in coffee branches non destructively. A linear estimation model was then built, to establish a correlation between the number of fruit samples automatically counted and the actual number observed in the field. The MVS enabled bot counting fruit and estimating their ripeness. Araújo (2015) developed an inspection system to assess bean quality and type. After acquiring bean images from a chamber on a conveyor, all pixels were assigned an intensity value according to the similarity between pixels and the typical intensity values of beans and the background. A correlation-based granulometry method was used to link to all bean grains their main features, such as eccentricity, size, and rotation angle. Later, K-means and KNN were adopted to classify beans into three categories: Carioca beans, Mulatto beans, and Black beans. The classification accuracy achieved was 99.88%.

Sanaeifar (2018) developed a non-destructive method which used dielectric spectroscopy and computer vision system to assess the quality of virgin olive oil. The correlation-based feature selection (CFS) method was used to extract dielectric and color features to reduce the dimension of input data. A classifier model selection was conducted using Artificial Neural Networks (ANN), SVM, and Bayesian networks (BN). Six parameters, including peroxide value (PV), p-Anisidine value (AV), total oxidation value (TOTOX), modeling free acidity (FA), Ultraviolet (UV) absorbance at 232 nm and 268 nm ($K_{232}$, $K_{268}$), chlorophyll and carotenoid, were used as quality indices of virgin olive oil and were predicted by the classifier models. The performance of BN outweighed ANN and SVM, with a 100% accuracy.

Deak (2015), Fernández-Espinosa (2015), Huang (2014), Xie (2014), Barreto (2018) proposed a similar approach to determine specific important contents in tomato juice, olive, soybean, sesame, and cheese, respectively. The authors used an MVS to capture the target's hyperspectral image in different spectral ranges, depending on the specific target, and subsequently corrected and segmented the image. Statistical algorithms such as Principal Component Regression (PCR), Successive Projections Algorithm (SPA), and PLSR were used to select the most significant wavelengths associated with the detection and prediction of certain compounds in the target food.

Song (2014) adopted a bag-of-words model to locate fruits in images and combined several images from different views to estimate the number of fruits with a novel statistical model. The bag-of-words model refers to a simplified expression model under natural language processing and information retrieval. With this model, text such as sentences or documents can be represented by a bag containing them. This representation method does not consider the grammar and the order of the words. Recently, the bag-of-words model has also been applied in the field of computer vision.

### *4.1.3. Traditional ML in Food Process Monitoring and Packaging*

Aghbashlo (2014) proposed using intelligent systems to monitor and control the drying speed of food, with the goal of improving the quality of the dried food. The proposed system inspected the texture, colour, size, and shape of the food by using co-occurrence, Fourier transform, wavelet transform, thresholding, and masking the input images. Subsequently, PCA and FCM were used to control the moisture values in the process, achieving optimal dryness. Liu (2016) solved the local homogeneous fragmentations or patches of complex grain images with a novel method. This method adopted multiscale and omnidirectional Gaussian Derivative Filtering (GDF), and a Sparse Multikernel-Least Squares Support Vector Machine (SMK-LSSVM) classifier. The authors proved that their method could be successfully used on an assembly line. Zareiforoush (2016) designed an intelligent system which used machine vision and fuzzy logic methods to control the rice whitening process. The intelligent system controlled the process by analyzing the degree of milling and the percentage of broken kernels. Subsequently, the output of the fuzzy logic algorithm (which was the whitening pressure level) could be used to realize the process monitoring. The authors proved that the system's efficiency was 31.3% higher than human labour, and the accuracy of the system's determination was 89.2%. Mogel (2014) proposed a segmentation method to calculate the region ratio between the browning parts on a cookie and the whole cookie, to monitor the baking process and inspect the quality of baked goods. Also, this method could be utilized to evaluate the safety of potato chips, because the browning ratio of potato chips has an excellent linear correlation ($R^2 > 0.88$) with acrylamide concentration.

### *4.1.4. Traditional ML in Foreign Object Detecting*

Lorente (2015) used a laser-light backscattering imaging system to detect the decay due to fungi in citrus fruit and achieved a maximum overall classification accuracy of 93.4%, using a Gaussian–Lorentzian cross product distribution function. Coelho (2016) adopted a binary decision tree to analyze and classify clam images captured by a transillumination technique. This online MVS was able to flatten the clam thickness and detect parasites in clams. A self-generated spatial reference system was used to classify the parasites, locate them and determine their shape. The grating-based X-ray imaging proposed by (Einarsdóttir et al., 2016) offers three modalities to detect objects (such as wood chips and insects) the textures of which cannot be detected by the classic X-ray method. Images of seven different food products containing foreign objects (such as hairs, staples, and metal pieces) were evaluated by determine the Mahalanobis distance from food models. The food models were created by fitting the food images alone or with texture features to a Gaussian Mixture Model (GMM), a linear combination of multiple Gaussian distribution functions. Einarsson (2017) proposed the Sparse Linear Discriminant Analysis (SDA) to detect foreign objects. SDA is a supervised statistical learning method and is a sparse version of the LDA (Clemmensen et al., 2011). They used two data sets, obtained using spring rolls and minced meat with foreign objects, and obtained X-ray images. Then the authors divided each image into several sub-regions, and SDA was used to classify regions containing contain foreign objects and regions which did not contain them. Dutta (2015) proposed a novel non-destructive technique to detect a neurotoxin substance called acrylamide from potato chips. This technique extracted the statistical features from the images in the spatial domain, and then used an SVM classifier to predict the presence of acrylamide in the food. The accuracy and sensitivity of the predictions were 94% and 96%, respectively. The summary of the applications of MVS with traditional machine learning methods in food processing is shown in Table 4.

**Table 4** Applications of MVS with traditional ML in food processing.

| Products | Species | Application | Classification Methods | Evaluation | Reference |
|---|---|---|---|---|---|
| Animal | Beef | Predicting | a regression model | $R^2_{adjusted}$ = 98.2, $P < 0.05$ | (Amani et al., 2015) |
| | Clam | Detecting | Binary DT | accuracy = 98% | (Coelho et al., 2016) |
| | Chicken | Grading | SPA–PLSR | $R_p$ = 0.801, $RMSEP$ = 0.157 | (Xiong et al., 2015) |
| | | Grading | PLSR | RMSEp, multiple results | (Yang et al., 2018) |
| | Egg | Grading | SPA-SVR, SVC | 96.3% for scattered yolk | (Zhang et al., 2015) |
| | Salmon | Grading | PLSR | $r_{cv}$ = 0.834 (*driploss*) $r_{cv}$ = 0.877 (*PH*) | (He et al., 2014) |
| | | Grading | TreeBagger | accuracy = 97.8% | (Xu et al., 2016) |
| Fruit | Apple | Grading | RVM | accuracy = 95.63% | (Zhang et al., 2014) |
| | | Grading | PLS, CARS | $r_p$ = 0.977, 0.977, 0.955 (three positions) | (Fan et al., 2016a) |
| | | Grading | MLR | $R$ = 0.90, $RMSECV$ = 6.99$N$ | (Sun et al., 2016) |
| | | Grading | CPLS | $r$ = 0.9327 | (Fan et al., 2016b) |
| | | Grading | a bi-layer model | $r$ = 0.9560 | (Tian et al., 2017) |
| | | Grading | PLS | $R^2_p$ = 0.83 | (Khatiwada et al., 2016) |
| | | Grading | PLS-DA, PBR | accuracy = 98% | (Keresztes et al., 2016) |
| | Apricot | Grading | PLS | | (Büyükcan et al., 2016) |
| | Blueberry | Grading | CARS-LS-SVM | accuracy = 93.3% (for healthy), accuracy = 98.0% (for bruised) | (Fan et al., 2017) |
| | | Grading | logistic function tree | accuracy = 95.2% | (Hu et al., 2016) |
| | | Grading | SVM | accuracy = 97% | (Leiva-Valenzuela et al., 2013) |
| | Cherry | harvesting | Bayesian | accuracy = 89.6% | (Amatya et al., 2015) |
| | Citrus | Detecting | Gaussian–Lorentzian | accuracy = 93.4% | (Lorente et al., 2015) |
| | Mango | Grading | SVR, MADM | accuracy = 87%. | (Nandi et al., 2016) |
| | | Grading | Fuzzy classifier | accuracy = 89% | (Naik et al., 2017) |
| | Peach | Grading | SPA | accuracy = 100% | (Sun et al., 2017) |

| | | Grading | An improved watershed segmentation algorithm | accuracy = 96.5% (for bruised), accuracy = 97.5% (for sound) | (Li et al., 2018) |
|---|---|---|---|---|---|
| | Pear | Grading | SPA-SVM | accuracy = 93.3%, 96.7% | (Hu et al., 2017) |
| | Pomegranate | Grading | PLS | $r = 0.97$ | (Khodabakhshian et al., 2016) |
| | Strawberry | Grading | SVM | accuracy = 100% | (Liu et al., 2014) |
| Vegetable | Tomato | Grading | DSSAEs | accuracy = 95.5% | (Iraji, 2018) |
| | Onion | Grading | SVMs | accuracy = 88.9% | (Wang et al., 2015) |
| | Potato | Grading | LDA-MD for color | above 90% for 5 potato cultivars (color) | (Noordam et al., 2000) |
| Others | Beans | Classifying | K-means and KNN | accuracy = 99.88% | (Araújo et al., 2015) |
| | Cheese | Grading | PLSR | $R^2 = 0.8321$ | (Barreto et al., 2018) |
| | Coffee bean | Grading | linear estimation models | $R^2 = 0.93$ | (Ramos et al., 2017) |
| | Cookie, Potato Chips | Monitoring | non-destructive computer vision-based image analysis | $R^2 = 0.895$ | (Ataç Mogol et al., 2014) |
| | Dried food | Monitoring | PCA, FCM | | (Aghbashlo et al., 2014) |
| | General | Detecting | GMM | multiple results | (Einarsdóttir et al., 2016) |
| | Grain | Monitoring | SMK–LSSVM | accuracy = 98.13% | (Liu et al., 2016) |
| | Olive | Grading | PLSR, PCA, LDA | multiple results | (Fernández-Espinosa, 2015) |
| | Olive oil | Grading | ANN, SVM, BN | accuracy = 100% with BN | (Sanaeifar et al., 2018) |
| | Potato Chips | Detecting | SVM | accuracy = 94% | (Dutta et al., 2015) |
| | Rice | Monitoring | Fuzzy logic | accuracy = 89.2% | (Zareiforoush et al., 2016) |
| | Sesame | Grading | CARS-LS-SVM, CARS-LDA | accuracy = 100% | (Xie et al., 2014) |
| | Soybean | Grading | PLSR | multiple results | (Huang et al., 2014) |
| | Spring rolls, minced meat | Detection | SDA | 5% error with 10-fold cross-validation | (Einarsson et al., 2017) |
| | Tomato Juice | Grading | PLSR | $R^2 = 0.75$ | (Deak et al., 2015) |
| | Walnut | Grading | SVM | | (Tran et al., 2017) |

## 4.2. Deep Learning Methods

Deep learning, also known as a deep neural network, is a learning method for building a deep architecture and generating a model by iterating functions in multiple layers. This method was proposed by in (Hinton et al., 2006). The learning effectiveness of deep learning is superior to traditional machine learning methods, although the interpretability is not as good as traditional learning methods.

### 4.2.1. Algorithms

Deep learning algorithms are based on representational learning of data. Observations (such as images) can be expressed in various ways, such as a vector of intensity values for each pixel, or more abstractly as a series of edges, or regions of a particular shape, as an example. There have been several deep learning frameworks, such as Convolutional Neural Networks, Recurrent Neural Networks, and Fully Convolutional Networks. They have been applied in computer vision, speech recognition, natural language processing, and bioinformatics and have achieved excellent results.

***Artificial neural networks (ANN).*** ANN is a relatively simple structure. It includes a simple perception, which uses mathematical functions to find the weighted sum of inputs and outputs its results. A typical structure of ANN is shown in Fig. 10. The nodes in the graph are called neurons. There are three layers in this graph. The first layer is the input layer, the second layer is the hidden layer, and the third layer is the output layer. The input layer accepts the input of the external world and converts it to the pixel intensity of the image or other features that can be quantified. The output layer is used to obtain probability prediction results (Zurada, 1992).

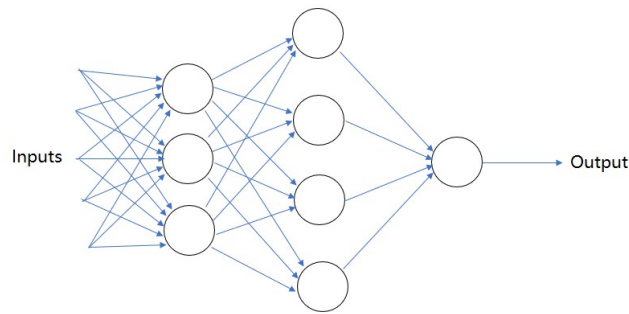

**Figure 10** An example structure of ANN.

***Back-Propagation (BP).*** Back Propagation (BP) network is a multi-layer feed-forward network trained by the error back-propagation algorithm and is one of the most widely used neural network models. This method calculates the gradient of the loss function for the weight in the network (Hecht-Nielsen, 1992). The gradient is fed back to the optimization algorithm to update the weights to minimize the loss function. Back-propagation requires a known output expected for each input value, in order to calculate the gradient of the loss function. The BP algorithm's learning process consists of forward propagation and backpropagation to perform repeated iterations of incentive propagation and weight update, as shown in Fig. 11. In the forward propagation process, the input information is processed layer by layer and passed to the output layer when passing through the hidden layers. Subsequently, the partial derivative of the objective function to each neuron's weights is calculated layer by layer, to form the ladder between the objective function and the weight vector, with the purpose of optimizing the weight. The learning is completed during the weight modification process. When the error reaches the expected value, the network learning ends.

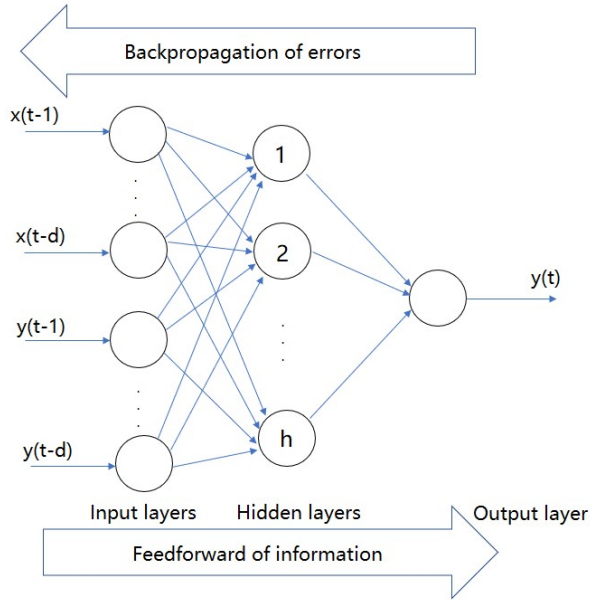

**Figure 11** An example structure of BP network.

***Convolutional Neural Networks (CNN).*** CNN includes a series of convolutional layers, nonlinear layers, pooling (down-sampling) layers, fully connected layers, and finally, the output (Krizhevsky et al., 2012). The output of CNN is the probability that best describes a single category or group of categories of the image. A CNN model of image recognition is shown in Fig. 12. The input image can be interpreted as several matrices, and the output is the determination of what object this picture is most likely to be. In the convolutional layer, the dot product between the input image and the weight matrix of the filter is calculated. The result is used as the output of the layer. The filter will slide across the entire image and repeat the same dot product operation. The process of convolution can be expressed as given in (5):

$$s(i,j) = (X*W)(i,j) + b = \sum_{k=1}^{n\_in}(X_k * W_k(i,j) + b) \tag{5}$$

where $n\_in$ is the number of input matrices or last dimension of the tensor, $X_k$ represents the $k^{th}$ input matrix, $W_k$ is the $k^{th}$ sub-convolution kernel matrix of the convolution kernel, and $s(i,j)$ is the value of the corresponding element of the output matrix related to the convolution kernel $W$. For the output of nonlinear layers, the ReLU activation function $f(x) = \max(0, x)$ is generally used. The ReLU function returns a value of 0 for each negative value in the input image but returns the same value for each positive value in the input image.

The pooling layer compresses each sub-matrix of the input tensor, so that the input feature space dimension becomes smaller than in previous layers, while the depth is not reduced. Depending on the task and on the characteristics of the model structure, the combination of the convolutional layer and the pooling layer can appear multiple times in the hidden layer. After several convolutional layers and pooling layers, a Fully Connected layer (FC) is obtained. In the FC layer, the last convolutional layer's output is flattened and connected to each node of the current layer with another node of the next layer. After obtaining an FC layer, the output layer uses the Softmax activation function to classify the image. The Softmax activation function can compress a K-dimensional vector $Z$ containing any real number into another K-dimensional real vector σ($Z$), so that the range of each element is between (0,1), and the sum of all elements is 1. This function can be expressed as given in (6):

$$\sigma(Z) = \frac{e^{z_j}}{\sum_{k=1}^{K} e^{z_k}} \; for \; j = 1, \dots, K \tag{6}$$

Another prevailing architecture called Recurrent Neuron Network (RNN) is also a typical model as CNN. CNN is often applied to spatially distributed data, while RNN introduces memory and feedback into neural networks and is often applied to temporally distributed data (Litjens et al., 2017).

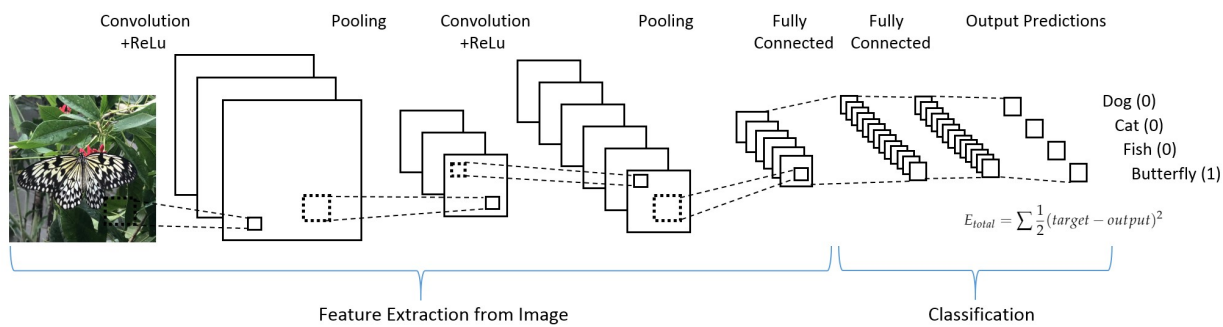

**Figure 12** The architecture of a typical CNN.

***Fully Convolutional Networks (FCN).*** Dissimilar to CNN, FCN replace the last fully connected layer of CNN with a convolutional layer, and the output is a labelled picture (Long et al., 2015). Unlike the classic CNN that uses fully connected layers to obtain fixed-length feature vectors for classification, FCN can accept input images of any size and up-sample the feature map of the last convolutional layer, using a deconvolution layer. The deconvolution layer can restore the output of the last layer to the same size of the input image, to generate a prediction for each pixel. Simultaneously, the spatial information in the original input image is retained, and finally, pixel-by-pixel classification is performed on the up-sampled feature map. FCN classifies images at the pixel level, thus solving the problem of semantic segmentation.

### *4.2.2. Deep Learning in Food Safety and Quality Evaluation*
Pan (2016) utilized a technique based on the hyperspectral imaging method to detect cold injury of peach, and utilized ANN to predict quality parameters. Poonnoy (2014) adopted ANN to classify the shapes of boiled shrimps based on the Relative Internal Distance (RID) values. The four shapes considered included 'regular', 'no tails', 'one tail', and 'broken body', and the RIDs were calculated by segmenting the shrimp images and drawing the co-related lines on the segmented contour. The overall prediction accuracy of the ANN classifier was 99.80%. Shafiee (2014) used MVS to capture and perform the colour transformation of the images of honey, while they used the ANN to predict key quality indices of honey (ash content, AC, antioxidant activity, AA, and total phenolic content, TPC). Zareiforoush (2015) proposed a computer vision and metaheuristic techniques to classify milled rice grains. Their method extracted the shape and size features, texture features, and colour features to build a primary feature vector, and selected the most important features based on the "greedy hill-climbing" and backtracking algorithm. The final feature vector was then used to train the ANN, SVM, Decision Trees (DT), and BN independently. The results indicated that the deep learning architecture – ANN was the best classifier.

Wan (2018) designed an MVS to acquire tomato images, and the ROIs were segmented from the whole images. Finally, a Back-Propagation Neural Network (BPNN) was used to classify the maturity level of Roma tomato and pear tomato. Huang (2016) used computer vision and NIR spectroscopy methods to obtain information about the organoleptic and structural changes of fish based on fish images. The PCA was employed to extract the most critical features from the data set, and a BackPropagation Artificial Neural Network (BP-ANN) was built to predict the fish freshness by training the algorithm to assess the extracted features. Khulal (2016) compared ACO (Ant Colony

Optimization)-BPANN to PCA-BPANN when extracting six textural features and predict chicken quality by quantifying the total volatile essential nitrogen content in chicken. The results showed that the performance of ACO-BPANN achieved RMSEp = 6.3834 mg/100g and R = 0.7542, which surpassed the performance of PCA-BPANN. Zhu (2020) developed a two-layer system to grade bananas and detect the defective spots on banana peels, with an overall accuracy of 96.4%. The first layer in their system is a SVM classifier and the second layer is a YOLOv3 (You Only Look Once Version 3) model. The YOLO model is a real-time object detection model, which is capable of detecting and locating small object locations with a high speed.

Pu (2015) obtained pork images via a hyperspectral imaging system in the 400 - 1000 nm VIS-NIR range and calibrated the images with dark and white reference images. Then, the pork muscle was segmented from the background. A variable selection method called Uninformative Variable Elimination and Successive Projections Algorithm (UVE-SPA) was adopted to find the major wavelengths with co-relations with raw spectral patterns pork categories. Histogram Statistics (HS), Gray Level Co-occurrence Matrix (GLCM), and Gray Level-Gradient Co-occurrence Matrix (GLGCM) were used to extract textural features separately, and Probabilistic Neural Network (PNN) was trained to predict the fresh and frozen-thawed level of the pork. PNN is a branch of radial basis network, which is a form of a feedforward network. It belongs to a supervised network classifier based on the Bayesian minimum risk criterion. Wu (2018) also developed a method to grade rice. The authors developed a Deep-Rice system by modifying the softmax loss function in a multi-view CNN and applied the CNN model to extract the discriminative features from several angles of the rice. Additionally, the authors built a data set of rice – FIST-Rice. With 30,000 rice kernel samples, this data set can be used to research food security.

### *4.2.3. Deep Learning in Food Process Monitoring and Packaging*

Teimouri (2018) extracted colour, geometrical aspects, and texture features with a CCD camera from chicken portions, then adopted the PLSR, LDA, and ANN to classify the data to achieve on-line separation and sorting of the food product. The conveyor that carried the food moved at 0.2 m/s, and it took 15 ms to process one image. The system's overall accuracy was 93%, and 2800 food samples could be sorted in one hour. De Sousa Ribeiro (2018a) offered an adaptable CNN-based system to identify the missing information on food packaging labels. In this research, K-means clustering and KNN classification algorithms were used to extract the data's centroids and provided them to CNN. Also, De Sousa Ribeiro (2018b) proposed an end-to-end deep neural network for detecting the dates on food packages. Optical Character Verification (OCV) and Recognition (OCR) (which are methods to convert printed characters into computer text) were used to extract the ROI and feed the data to a Fully Convolutional Network (FCN) to locate the data. Then a Maximally Stable Extremal Regions (MSER) was adopted to segment the data. FCN solves semantic segmentation by pixel-level classification of images, and MSER is used for spot detection in an image.

### *4.2.4. Deep Learning in Foreign Object Detecting*

Pujari (2013) proposed a system to detect and classify fungi contamination on commercial crops. The authors applied Discrete Wavelet Transform (DWT) and PCA to extract significant features from crops images and feed them to Mahalanobis distance and PNN classifiers. The final prediction accuracy of these two classifiers was 83.17% and 86.48%, respectively. Ma (2017) carried out a method to segment greenhouse vegetables with foliar disease spots in the real field. The authors adopted Comprehensive Color Feature (CCF), including colour information Excess Red Index (ExR), H component of HSV colour space and b* component of L*a*b* colour space, as well as region growing, to segment disease spots from clutter backgrounds. This method overcame the issues of uneven illumination and clutter background encountered in the real field. The CCF were then sent to a CNN architecture to classify the spots. The precision of this algorithm achieved an accuracy of 97.29%. Dos Santos Ferreira (2017) developed software to segment and detect weeds growing among soybean crops. This software classification function is based on a CNN trained with crop images captured in real fields by three professional drones. The system adopted the superpixel segmentation algorithm - Simple Linear Iterative Clustering (SLIC), which groups the pixels into atomic regions to replace the pixel grid, and to segment the undesirable weeds from soil and soybean crops. Ebrahimi (2014) proposed a system to distinguish wheat grains from weed seeds, with the goal of determining the purity and grade of the wheat seeds. The authors used a hybrid of Imperialist Competitive Algorithm (ICA) (which is used for optimization problems) and ANN as classifier, to identify the optimal characteristic parameters. Al-Sarayreh (2019) proposed a foreign object detection system for meat products. This system used real-time hyperspectral imaging to extract both spectral and spatial features of the target food, integrating it with a sequential deep-learning framework that is composed of region-proposal networks (RPNs) and 3D-CNNs. Rong (2019) developed two CNNs. These two models can detect foreign objects such as flesh leaf debris and metal parts in walnuts and segment the foreign objects. The accuracy of segmentation and classification are 99.5% and 95%, respectively.

Table 5 summarizes recent deep learning methods applied in the field of food processing.

**Table 5** Applications of MVS with Deep Learning in food processing.

| Products | Species | Application | Classification Methods | Evaluation | Reference |
|---|---|---|---|---|---|
| Animal | Chicken | Monitoring | PLSR, LDA, ANN | accuracy = 93% | (Teimouri et al., 2018) |
| | | Grading | ACO-BPANN | RMSEp = 6.3834 mg/100g, R = 0.7542 | (Khulal et al., 2016) |
| | Cod Fillets | Grading | SVM, CNN | accuracy = 100% with CNN | (Misimi et al., 2017) |
| | Fish | Grading | PCA, BP-ANN | accuracy = 93.33% | (Huang et al., 2016) |
| | Meat | Detecting | RPN plus CNN | accuracy = 81.0% | (Al-Sarayreh et al., 2019) |
| | Pork | Grading | GLCM, GLGCM, PNN | accuracy = 92.02% | (Pu et al., 2015) |
| | Shrimp | Grading | ANN | accuracy = 99.80% | (Poonnoy et al., 2014) |
| | | Monitoring | MLP-ANN | $r > 0.99$ | (Hosseinpour et al., 2015) |
| | Squid | Classifying | Improved faster R-CNN | accuracy > 0.8 | (Hu et al., 2019) |
| Fruit | Banana | Grading Detecting | SVM, YOLOv3 | accuracy = 96.4% | (Zhu et al., 2020) |
| | Peach | Grading | ANN | accuracy = 95.8% | (Pan et al., 2016b) |
| | Tomato | Grading | BPNN | accuracy = 99.31%; | (Wan et al., 2018) |
| Vegetable | General | Detecting | CCF, CNN | accuracy = 97.29% | (Ma et al., 2017) |
| | Almond | Grading | SEM | mean error = 0.63% (for mass) | (Vidyarthi et al., 2020) |
| | Crop | Detecting | DWT, PCA, PNN | accuracy = 86.48% | (Pujari et al., 2013) |
| | Fluid | Monitoring | CNN | accuracy = 99.76% | (Katyal et al., 2019) |
| | General | Packaging | OCV, OCR, FCN, MSER | accuracy > 97.1%% | (De Sousa Ribeiro et al., 2018b) |
| | Grain | Detecting | ICA-ANN | multiple results | (Ebrahimi et al., 2014) |
| | Honey | Grading | ANN | multiple results | (Shafiee et al., 2014) |
| Others | Jujube | Grading | PCA, SVM, LR, CNN | accuracy > 85% | (Feng et al., 2019) |
| | Oil Palm | Grading | PCA, GANN | accuracy = 84.5% | (Silalahi et al., 2016) |
| | Pistachio | Grading | PCA, AlexNet, GoogleNet | accuracy = 99% with GoogleNet | (Farazi et al., 2017) |

|  | Retail Foods | Packaging | K-means, CNN | accuracy = 76.4% (dataset 1) accuracy = 97.1% (dataset 2) | (De Sousa Ribeiro et al., 2018a) |
|  |  | Grading | ANN, BN, DT, SVM | accuracy = 98.72% with ANN | (Zareiforoush et al., 2015) |
|  | Rice | Grading | multi-view CNN architecture | accuracy = 84.43% | (Wu et al., 2018) |
|  |  | Grading | CNNs | multiple results | (Desai et al., 2019) |
|  | Soybean | Detecting | SLIC, CNN | accuracy = 98% (broadleaf) accuracy > 99% (grass weeds) | (dos Santos Ferreira et al., 2017) |
|  | Walnuts | Detecting | CNNs | accuracy = 95% | (Rong et al., 2019) |

## 5. Open challenges and Future Trends

The previous sections discussed different MVS applications for food processing. Although MVS is most often successfully used in diverse scenarios, most technologies still have limitations and shortcomings. Moreover, due to the diversity and complexity of the food industry's research objectives and the MVS technology's characteristics, applications based on MVS technology still present the following problems:

1. MVS technology has high requirements for measurement conditions and the environment. For example, if the target's background image is too noisy and the illumination conditions are unsatisfactory, detection and classification may be insufficient. Food processing and production environments and applications are complex and variable, rendering necessary the use of different algorithms for different research objectives and environments. As a result, the robustness and reliability of MVS may not meet requirements.
2. Due to the variety of foods and the diversity of features, MVS still presents some deficiencies in detecting food information and extracting features. More accurate algorithms need to be developed to detect foods with small color changes or unusual shape features. Meanwhile, imaging technologies are largely unable to correlate the appearance of food to their smell, which is one of the essential reference features for food safety and quality evaluation. Therefore, it is difficult to use imaging technology for the online detection of smells.
3. Since the core of MVS is image processing, the storage and computation ability of large amounts of image data will increase the processing system's requirements. The image processing system's cost will increase significantly when multiple imaging technologies and inspection technologies need to be integrated, especially when an online inspection is performed.
4. MVS, which currently includes a decision execution module such as a robot arm, is not mature enough. As a result, it cannot be used for large-scale food production applications and it cannot be fully intelligent. Moreover, due to the mechanical control system's limitations, MVS still cannot meet the requirements on some occasions with high real-time requirements.

Due to the complexity and long-term nature of the problems, the applications of MVS in the food industry will continue to develop for some time, especially when the goal is to carry out large-scale production applications. Future research and development direction should mainly focus on the following aspects:

1. Image processing, which is the core of MVS. Improving existing algorithms or developing more efficient algorithms to improve the processing efficiency and robustness of MVS is still a vital prerequisite for future machine vision applications and development.

2. Embedded vision systems, which are the inevitable scope for the development of MVS in the next few years. Because of the compact structure, fast processing speed, and low cost of embedded systems, MVS applications will be popularized on a large scale more rapidly.
3. MVS that incorporates multiple technologies, which is already an active research area and should be further studied in the future. Especially now that the Internet of Things technology is prevalent, MVS needs more user-side interface support to form smart systems, thereby realizing smart agriculture. Technologies such as edge computing in the Internet of Things can also enable the image processing and image interpretation steps in machine vision, to use fewer resources and achieve less latency.

## 6. Conclusion

This survey helps readers understand the principles of machine vision systems, the role of each step of image processing, and the leading development and applications of machine vision systems in the field of food processing in the recent five years. MVS is widely used in food safety inspection, food processing monitoring, foreign object detection, and other domains. It provides researchers and the industry with faster and more efficient working methods and makes it possible for consumers to obtain safer food. The systems' processing capacity can be boosted to a large extent, especially when using machine learning methods. Although challenges remain, implementing machine vision systems in food processing is a general trend, and we expect an increased use of machine vision systems for food processing.